\documentclass[lettersize,journal]{IEEEtran}
\usepackage{amsmath,graphicx}
\usepackage{amssymb}
\usepackage{color,hyperref}
\usepackage[table,xcdraw]{xcolor}

\usepackage{subcaption}
\usepackage{times}

\usepackage{algorithm}
\usepackage{algorithmicx}
\usepackage{algpseudocode}
\usepackage{multirow}
\usepackage{booktabs} 
\usepackage{makecell} 
\usepackage{url}
\usepackage{bbm}  
\usepackage{bm}  
\usepackage{verbatim}
\usepackage{breakurl}
\usepackage{bbding}
\usepackage{tabu}

\usepackage{graphicx}
\usepackage{amsmath}
\usepackage{amssymb}
\usepackage{booktabs}

\begin{document}
\title{Growing Instance Mask on Leaf}

\author{Chuang~Yang,
	Haozhao~Ma,
	and~Qi~Wang*,~\IEEEmembership{Senior Member,~IEEE}
	
}



\maketitle

\begin{abstract}
	Contour-based instance segmentation methods include one-stage and multi-stage schemes. These approaches achieve remarkable performance. However, they have to define plenty of points to segment precise masks, which leads to high complexity. We follow this issue and present a single-shot method, called \textbf{VeinMask}, for achieving competitive performance in low design complexity. Concretely, we observe that the leaf locates coarse margins via major veins and grows minor veins to refine twisty parts, which makes it possible to cover any objects accurately. Meanwhile, major and minor veins share the same growth mode, which avoids modeling them separately and ensures model simplicity. Considering the superiorities above, we propose VeinMask to formulate the instance segmentation problem as the simulation of the vein growth process and to predict the major and minor veins in polar coordinates. 
	
	Besides, centroidness is introduced for instance segmentation tasks to help suppress low-quality instances. Furthermore, a surroundings cross-correlation sensitive (SCCS) module is designed to enhance the feature expression by utilizing the surroundings of each pixel. Additionally, a Residual IoU (R-IoU) loss is formulated to supervise the regression tasks of major and minor veins effectively. Experiments demonstrate that VeinMask performs much better than other contour-based methods in low design complexity. Particularly, our method outperforms existing one-stage contour-based methods on the COCO dataset with almost half the design complexity.
\end{abstract}


\section{Introduction}
\label{Sec:Introduction}
\IEEEPARstart{I}{nstance} segmentation is one of the challenging computer vision tasks, which provides essential information for many intelligent applications (such as security monitoring and self-driving). With the rapid development of deep learning, instance segmentation has achieved great progress. 

Some researchers follow the intuitive idea and propose two-stage instance segmentation schemes. Typically examples like Mask R-CNN~\cite{he2017mask}, Cascade R-CNN~\cite{cai2018cascade}, and PANet~\cite{liu2018path}. They formulate the instance segmentation problem as the combination of previous techniques (object detection and semantic segmentation). However, modules (such as RPN~\cite{ren2015faster} and ROIAlign~\cite{he2017mask}) restrict the training and inference efficiency deeply, which leads to expensive hardware resources and time costs for further research. Some works~\cite{bolya2019yolact,chen2020blendmask,wang2020solo} follow this issue and construct one-stage~\cite{redmon2016you,tian2019fcos} pipelines, while the hybrid design of detection and segmentation makes it hard to assemble instance masks.

Recent trends~\cite{xie2020polarmask,zhang2022e2ec} aim to represent instances through dense contour points and predict them via detection techniques only, which ensures simple pipelines. However, to cover instances precisely, they have to define a \textbf{large} and \textbf{fixed} number of contour points, which suffers from the following problems: (1) \textit{A large number of points increases the model complexity.} For example, to gain a strong ability to cover masks, PolarMask~\cite{xie2020polarmask} has to increase the directions in polar coordinates and to encode each of them by CNN layers, which complicates the model. (2) \textit{The fixed number of points are hard to fit instances with different shapes, which accelerates the increase of model complexity.} For example, E2EC~\cite{zhang2022e2ec} decomposes instances into multiple smooth and twisty parts and covers them through a fixed number of points. They have to define more points to cover twisty parts, while it leads to heavily redundant points for smooth parts. The imbalance between smooth and twisty parts accelerates the increase of model complexity.

\begin{figure}
	\centering
	\subcaptionbox{Details of VeinMask structure}{\includegraphics[width=.9\linewidth]{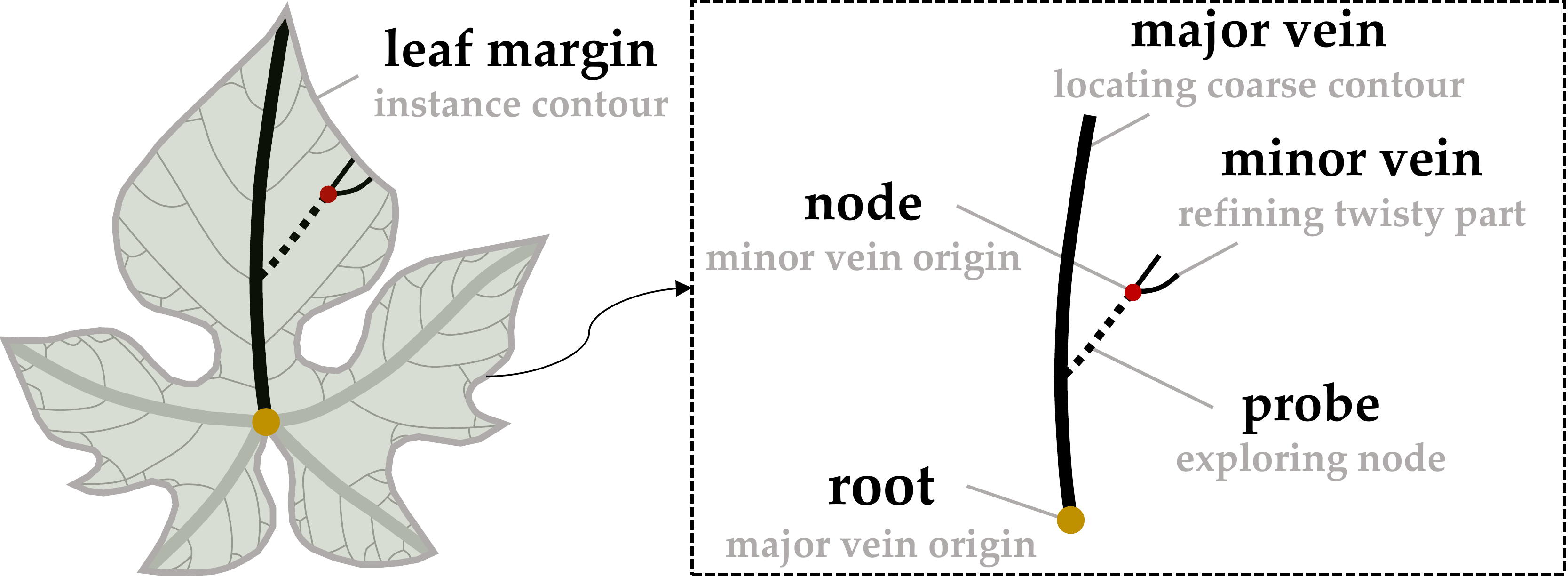}}	
	\subcaptionbox{PolarMask~\cite{xie2020polarmask}}{\includegraphics[width=.3\linewidth]{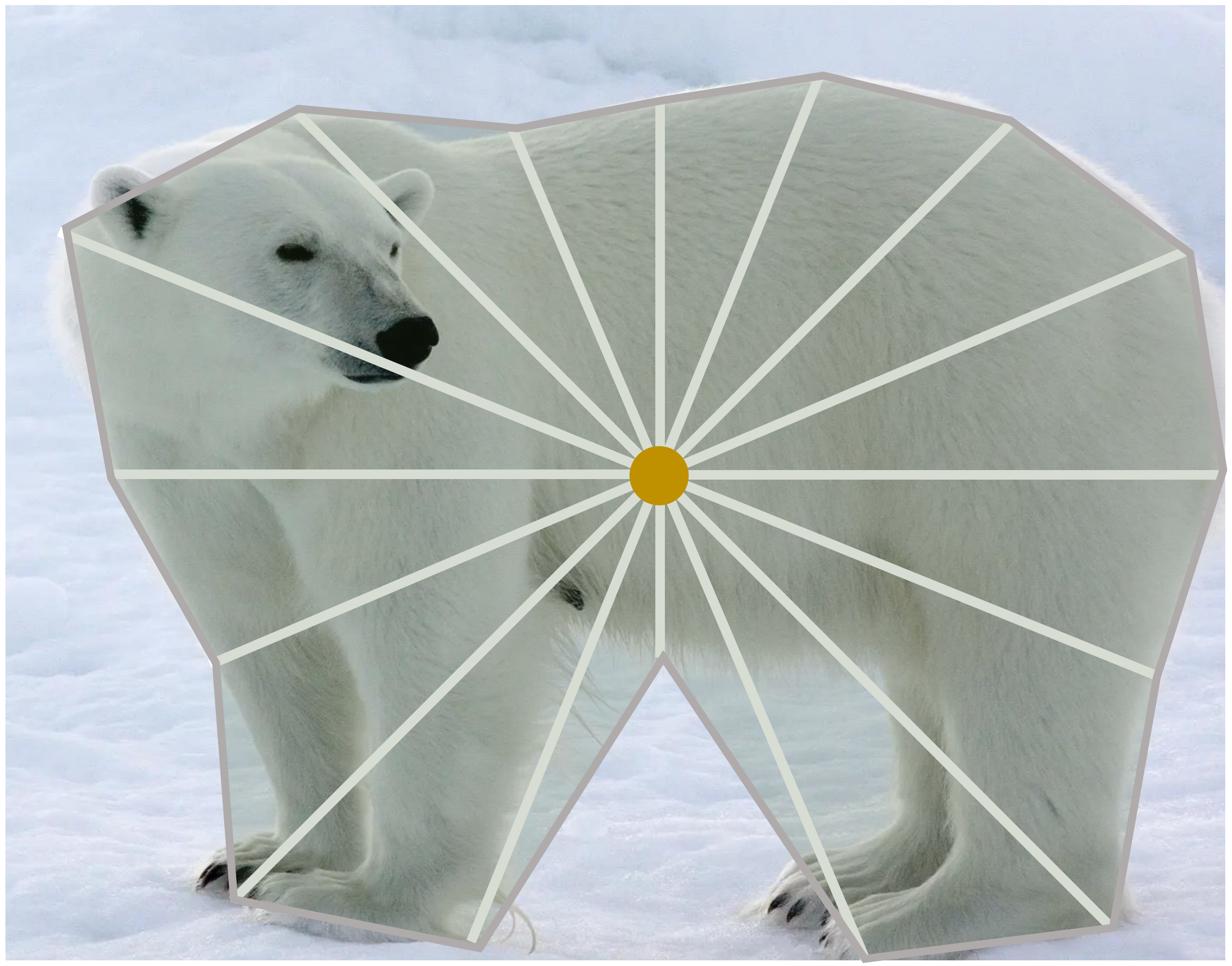}}
	\subcaptionbox{E2EC~\cite{zhang2022e2ec}}{\includegraphics[width=.3\linewidth]{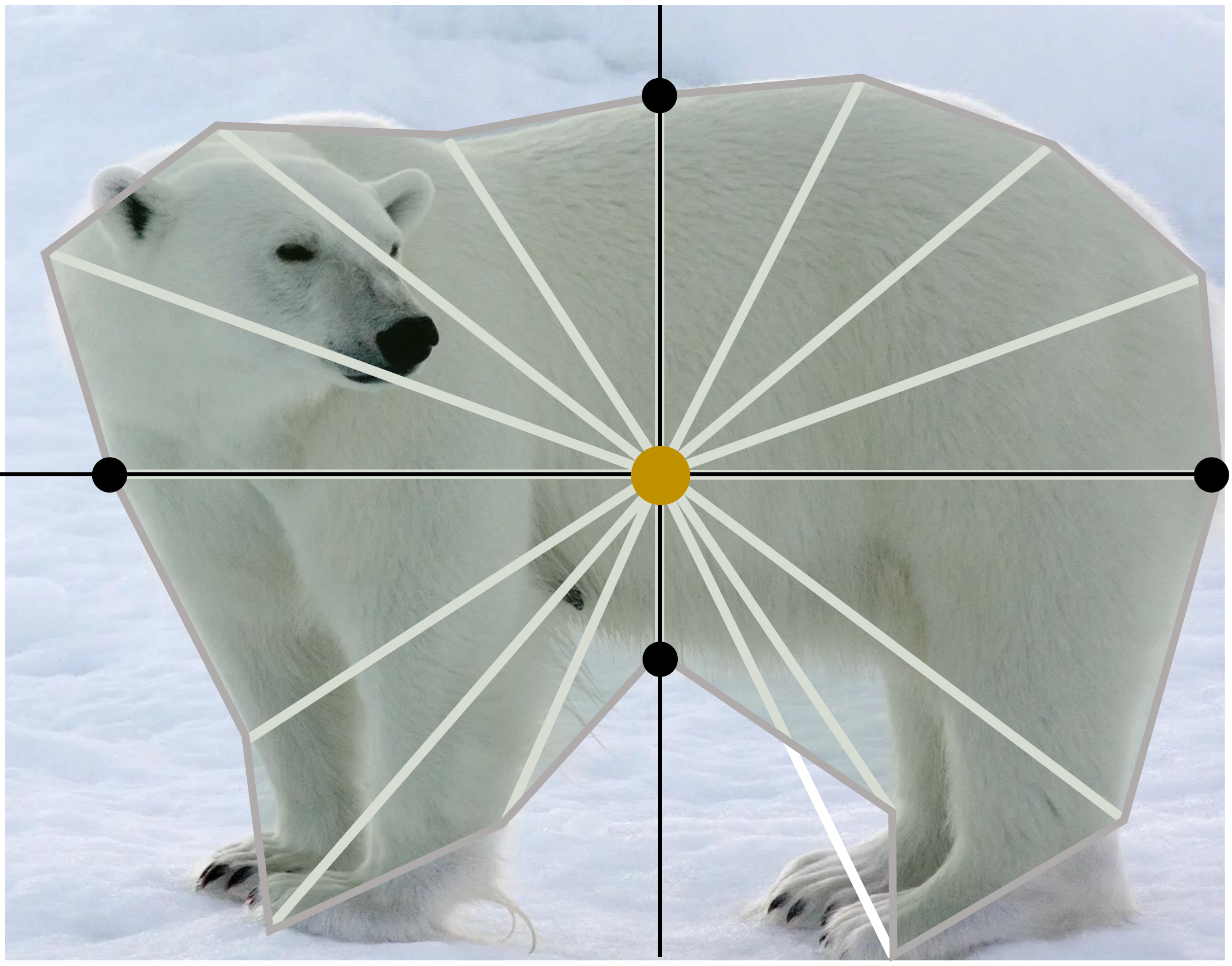}}
	\subcaptionbox{VeinMask (Ours)}{\includegraphics[width=.3\linewidth]{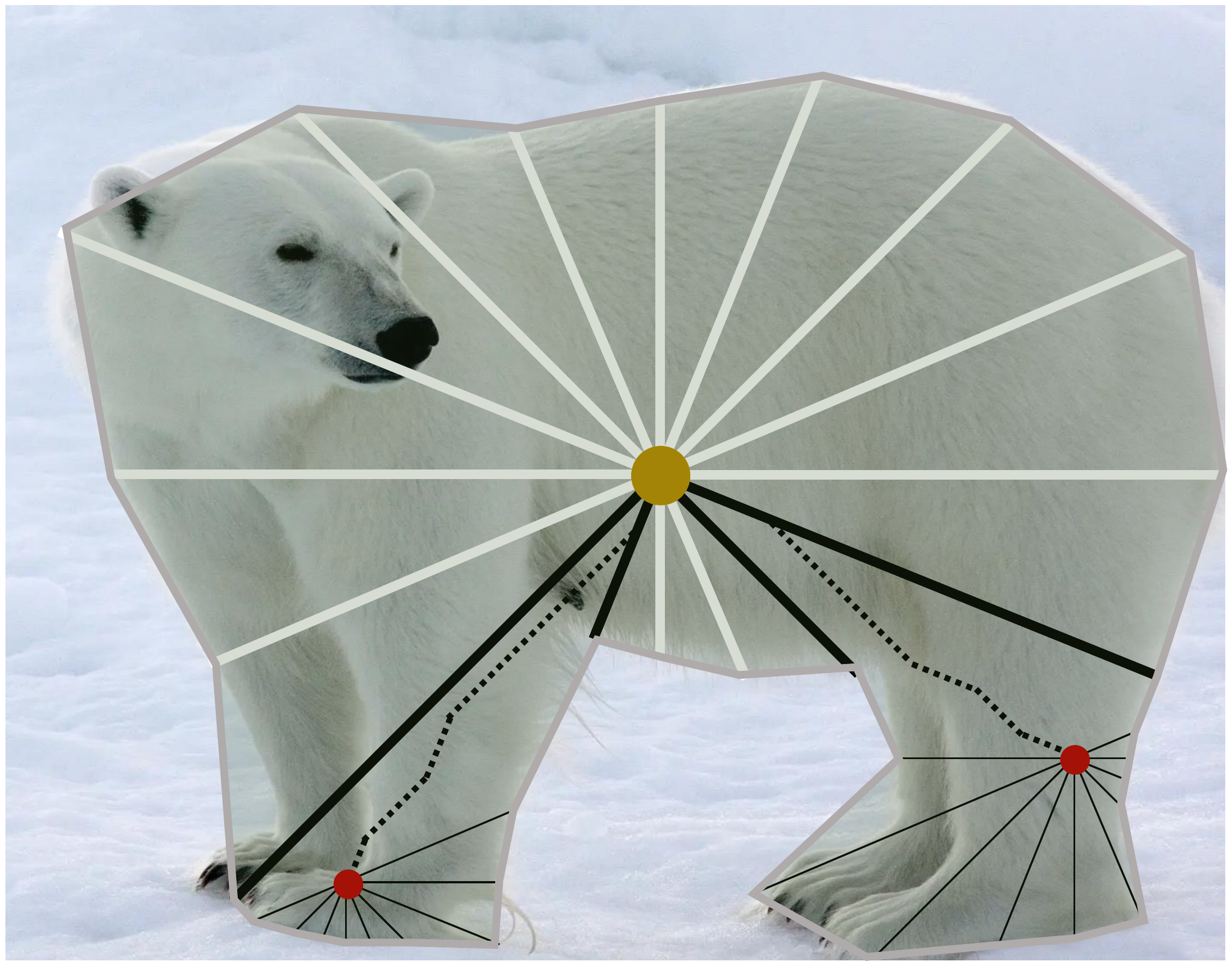}}
	\caption{Visualization of ideal segmented masks in low design complexity. The `\textbf{white lines}' in (b)--(d) are the lines between the instance centroid and predefined contour points. For better visualization, we redraw some `\textbf{white lines}' as `\textbf{black lines}' in (d) to highlight the lines that are used for refining twisty parts. (a) shows the details of vein mask construction. (b) and (c) represent the masks with current one-stage and multi-stage schemes, respectively. (d) is the segmented result from the vein mask.}
	\label{V1}
\end{figure}

Considering the issues above, we aim for pursuing to cover instance masks more precise with a simpler model by achieving the following two targets: (1) \textit{generating a large number of points via low design complexity to depict masks precisely }; (2) \textit{dynamic allocating an appropriate number of points to smooth and twisty parts of instances, respectively}. We combine morphology with deep learning and propose to simulate the leaf vein growth process to segment instance masks. As shown in Fig.~\ref{V1}(a), the leaf vein, a directed graph that can represent any complex geometries, is composed of major and minor veins. In the growth process, major veins sprout out from the root to locate the coarse leaf margin at first. Then, minor veins grow from the node to refine the twisty part. We observe the process and find that major and minor veins share the same growth mode. It means that we can grow the two kinds of veins without modeling them separately, which makes it possible for generating a large number of points with low design complexity. Meanwhile, minor veins occur only if major veins are unable to cover the twisty parts, which promises an appropriate number of points for each smooth and twisty part.

\begin{figure}
	\centering
	\includegraphics[width=.43\textwidth]{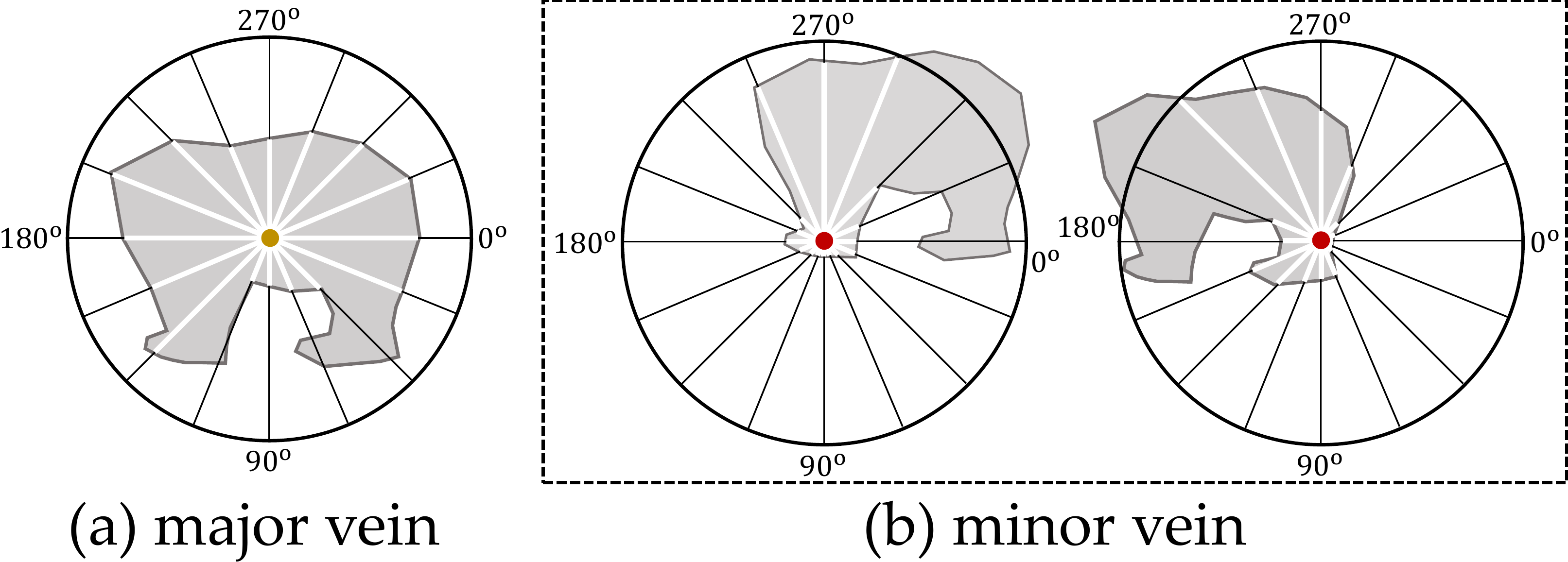}
	\caption{Visualization of major and minor veins built in the same polar coordinate with $n$ directions. `\textbf{orange circle}' and `\textbf{red circle}' are `\textbf{root}' and `\textbf{node}' respectively. `\textbf{while lines}' are veins.}
	\label{V2}
\end{figure}

Inspired by the superiorities above, we design VeinMask and construct a single-shot instance segmentation framework based on it, which simulates the leaf vein growth process to segment instances precisely with low design complexity. Concretely, we model major and minor veins through a polar coordinate with $n$ directions (as depicted in Figure~\ref{V2}) to assemble masks by the following steps: (1) locating the coarse contour through instance centroid (`root') classification and dense distance (`major vein') regression in the polar coordinate built at the root; (2) exploring the twisty part origin (`node') according to adjacent major veins; (3) refining the twisty part through dense distance (`minor vein') regression in the polar coordinate built at the node; (4) assembling instance contour by connecting all endpoints of major and minor veins in a clockwise direction.

Meanwhile, we follow the idea of centerness in FCOS~\cite{tian2019fcos} to present centroidness to help suppress low-quality instances in the inference process. Compared with the centerness in FCOS~\cite{tian2019fcos} and PolarMask~\cite{xie2020polarmask}, centroidness forces our method to focus on the instance centroid and spreads around in a Gaussian distribution, which is more effective for instance segmentation tasks and easier to learn for the model. Remarkably, the centroidness can be embedded in other one-stage regression-based instance segmentation methods seamlessly to bring performance gains. Besides, considering the vein plays a key role in assembling instance masks and the weak features make it hard to regress veins accurately, a surroundings cross-correlation sensitive (SCCS) module is proposed. It helps the model utilize the surrounding information of each pixel to encourage extracting strong expression features while ensuring the enhanced pixel dominance to the surroundings, which can suppress the negative effects brought by surrounding features. Furthermore, a Residual IoU (R-IoU) loss is formulated for supervising the regression of the major and minor veins. It inherits the IoU loss~\cite{yu2016unitbox} advantage to correlate the veins in all directions of polar coordinates. Particularly, different from the IoU loss, our R-IoU loss focuses on the residual between the predicted and real values, which can optimize the model more effectively. The contributions of this work are summarized as follows:

\begin{itemize}
	\item[1.] We combine morphology with deep learning to design a VeinMask, which simulates the leaf vein growth process to generate dense contour points for representing instances. It can achieve a stronger ability to cover masks precisely than previous contour-based methods with lower design complexity.	
	
	\item[2.] Centroidness is proposed for suppressing low-quality results. It focuses on the instance centroid and spreads around in a Gaussian distribution related to the instance geometries, which can easier bring significant performance gains for instance segmentation tasks compared to the centerness in FCOS and PolarMask.
	
	\item[3.] A surroundings cross-correlation sensitive (SCCS) module is introduced. It helps our model enhance the feature expression by utilizing the surrounding information of each pixel to encourage regression tasks. Importantly, the module ensures the enhanced pixel dominates to surroundings, which can suppress the negative effects brought by surrounding features.
	
	\item[4.] A Residual IoU (R-IoU) loss is formulated for supervising the regression of the major and minor veins. Remarkably, it inherits the IoU loss~\cite{yu2016unitbox} advantage and focuses on the residual between the predicted and real values, which is more effective for instance segmentation tasks compared to IoU loss and Polar IoU loss.

	\item[5.] We construct a single-shot instance segmentation framework based on VeinMask, which follows the design of one-stage bounding box object detection methods. It is simpler than mask-based methods both in the aspects of the CNN model and assembling masks process, and can achieve comparable performance with existing state-of-the-art (SOTA) methods.
\end{itemize}

\section{Related Work}
\label{Sec:Related Work}
\textbf{Mask-based instance segmentation methods.} The intuitive idea for instance segmentation is combining existing object detection and semantic segmentation techniques. Typically, Mask R-CNN~\cite{he2017mask} followed the design of Faster R-CNN~\cite{ren2015faster}. It predicted the bounding boxes of instances at first and then segmented precise masks within the boxes, which inspired plenty of following works (such as PANet~\cite{liu2018path}, Cascade Mask R-CNN~\cite{cai2018cascade}, and HTC~\cite{chen2019hybrid}). They improved Mask R-CNN in the aspects of feature extraction and R-CNN structure and achieved superior performance. However, these methods failed to segment large instances precisely because of ROIAlign operators. With the development of one-stage detection methods, some works~\cite{bolya2019yolact,chen2020blendmask,ying2021embedmask,tian2020conditional} proposed to segment masks on the whole feature maps directly instead of on the feature patches, which accelerated the inference speed while ensuring competitive performance. However, they had to assemble instance masks through a complicated process, which limited the model's performance. SOLO~\cite{wang2020solo,wang2020solov2} followed the design of YOLO~\cite{redmon2016you}. Different from classifying each pixel into a specific category, it segmented masks by predicting the relationships between each pixel and all grids. However, though SOLO~\cite{wang2020solo,wang2020solov2} enjoyed an efficient pipeline because of the grid mechanism, it made the detection of small instances difficult and depended on the settings of grids deeply.

\begin{figure*}
	\centering
	\includegraphics[width=.85\textwidth]{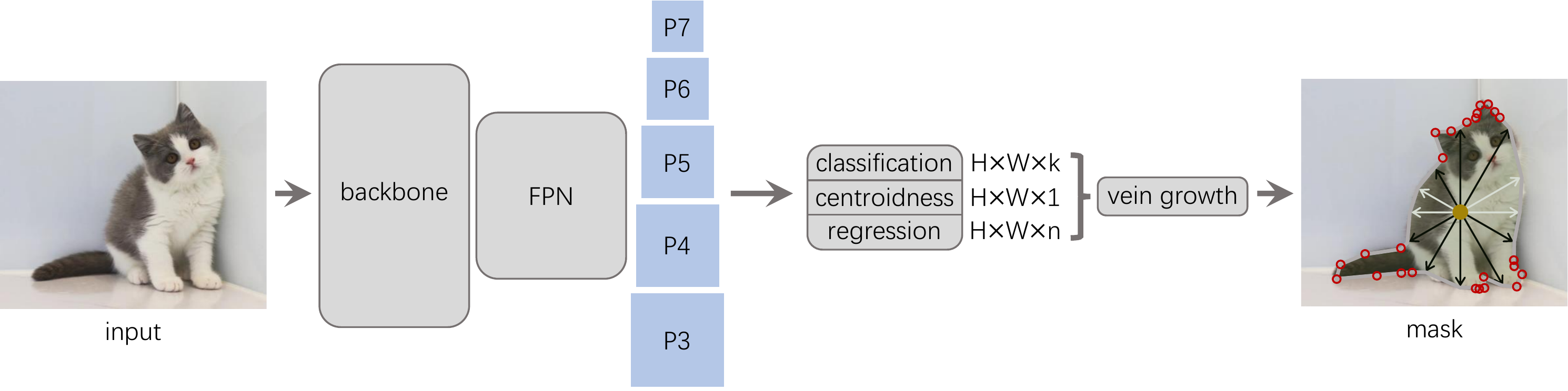}
	\caption{Overall architecture. The backbone and FPN are used to extract different level features. The classification, centroidness, and regression heads conduct on P3--P7 to generate three maps for each of them. $H$ and $W$ are the height and width of feature maps. $k$ and $n$ are the instance category number and direction number of the polar coordinate. Vein growth is responsible for assembling masks. The `\textbf{red circle}' in mask are the contour points of twisty parts. The `\textbf{white lines}' are the lines between the instance centroid and predefined contour points. We redraw some `\textbf{white lines}' as `\textbf{black lines}' to highlight the lines that are used for refining twisty parts.}
	\label{V3}
\end{figure*}

\textbf{Contour-based instance segmentation methods.} Mask-based methods adopt a hybrid design of detection and segmentation, which leads to inherent model complexity. To pursue a straightforward and effective instance segmentation pipeline, recent methods~\cite{yang2020dense,ling2019fast,liu2021dance,peng2020deep,wei2020point,zhang2022e2ec,xie2020polarmask,duan2021location} formulated instance segmentation problem as contour points prediction. DenseReppoints~\cite{yang2020dense} predicted a series of boundary points and the corresponding categories to represent instance masks. Some researches~\cite{ling2019fast,peng2020deep,wei2020point,liu2021dance,zhang2022e2ec} proposed to regress ordered contour point sequences. They improved the predicted point reliability through a multi-stage point refinement structure and achieved remarkable performance. But the refinement structure complicated the model seriously and led to expensive research costs (such as computational and time costs) for the following researchers. PolarMask~\cite{xie2020polarmask} inherited the design idea of anchor-free detection framework~\cite{tian2019fcos}. The authors regressed the offsets between instance centers and boundaries to generate dense ordered contour points for rebuilding masks, which could reach a quite fast detection speed that is almost equivalent to that of the one-stage detectors. However, the mask cover quality of PolarMask~\cite{xie2020polarmask} is limited.

\section{Methodology}
\label{Sec:Methodology}
We construct a single-shot framework based on VeinMask to segment instances precisely with low design complexity. Our method aims to simulate the leaf vein growth process to reconstruct instance masks. It assembles contours by the `root' and `node' classification and `major vein' and `minor vein' (see Figure~\ref{V1}~(a)) regression. Figure~\ref{V3} illustrates the details of our framework, which consists of backbone, FPN~\cite{lin2017feature}, three heads (including classification, centroidness, and regression heads), and vein growth (Figure~\ref{V1} shows the idea). They will be introduced next in detail.

\textbf{Building veins in polar coordinates.} As depicted in Figure~\ref{V1}~(a)(d), VeinMask represents instance masks according to the combinations of `root' and `major vein', and `node' and `minor vein' respectively. In practice, we define a single polar coordinate with $n$ directions to model the two combinations simultaneously (see Figure~\ref{V2}), where each direction in Figure~\ref{V2}~(a) and Figure~\ref{V2}~(b) will generate a contour point. It means there will be more than $n$ points by encoding a single polar coordinate, which makes it possible to cover masks precisely with low design complexity.

\textbf{Backbone and FPN.} Following previous instance segmentation methods, we extract hierarchical features (see Figure~\ref{V3}~$\mathrm{P_3}\sim\mathrm{P_7}$) for the detection of instances through the combination of backbone and FPN. Meanwhile, to meet the cases that multiple instances are close to each other, we follow FCOS~\cite{tian2019fcos} to allocate instances into different level feature maps from FPN. Concretely, supposing $l^*,t^*,r^*,b^*$ are the distances between centroids and instance bounding boxes in left, top, right, and bottom directions, the instance is distributed to $\mathrm{P_k}$ if $\mathrm{max}(l^*,t^*,r^*,b^*)\in\mathrm{S_k}$, where $\{\mathrm{S_k}|,k=3,4,5,6,7\}=\{(-1,64),(64,128),(128,256),(256,512),(512,+\infty)\}$.

\textbf{Classification head.} The classification head is responsible for locating instance centroids (the root in Figure~\ref{V1}~(a)) by classifying each pixel into $k$ categories, where $k$ is the number of instance categories in a specified dataset. It applies to all feature maps ($\mathrm{P_3}\sim\mathrm{P_7}$) from FPN and detects different sizes of instances on different levels of feature maps. We refer to FCOS~\cite{tian2019fcos} to construct the head through a few CNN layers. Different from FCOS adopts the midpoints of rectangular boxes as instance centers, we locate instances via centroids, which helps the proposed VeinMask to cover irregular-shaped masks more precisely.

\begin{figure}
	\centering
	\includegraphics[width=.41\textwidth]{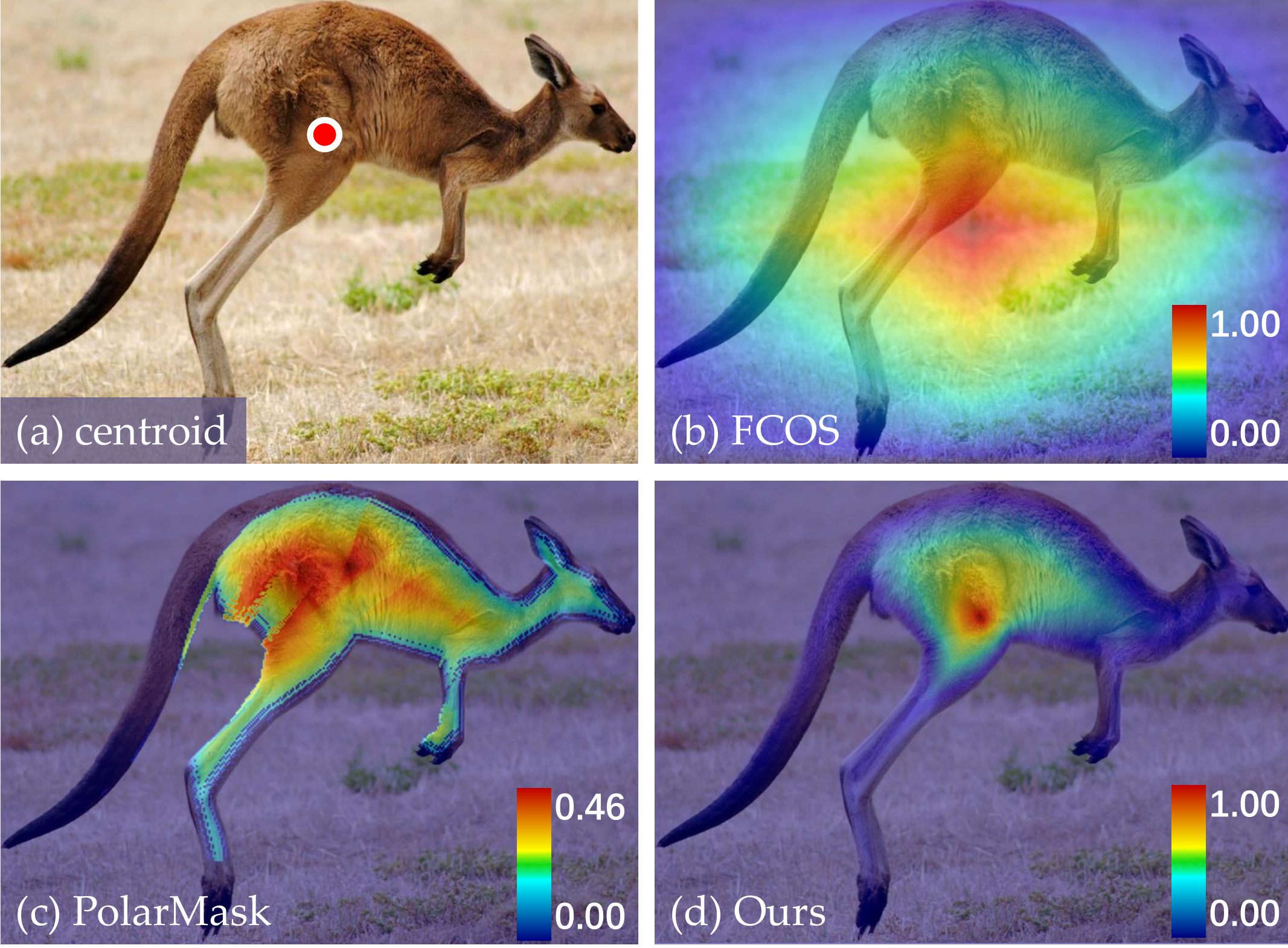}
	\caption{Visualization of centroidness. (a) shows the location of instance centroid. (b) and (c) are the centerness from FCOS and PolarMask, respectively. (d) illustrates our centroidness that focuses on the centroid and spreads around in a Gaussian distribution related to instance geometries.}
	\label{V4}
\end{figure}

\textbf{Centroidness head.} Centerness is introduced to suppress the low-quality results in FCOS~\cite{tian2019fcos} and PolarMask~\cite{tian2019fcos}. However, it may lead to information distortion that hinders the model from learning the correct centroid importance information in instance segmentation tasks. We follow the above issue and design centroidness for improving the reliability of the instance centroids. 

Concretely, FCOS assigns centerness a higher weight at the location that is closer to the rectangular box midpoint, which expresses the different importance between the center and around pixels accurately for rectangular boxes but not for instance masks. As shown in Figure~\ref{V4}~(b), the weight at the instance centroid location (see Figure~\ref{V4}~(a)~red circle) is smaller than around pixel weights. Besides, the worse thing is that the weight even may approximate 1 within the background region. The two factors hinder the model to learn centroid importance correctly to filter low-quality instances. PolarMask defines the centerness as the ratio of the $d_{min}$ to $d_{max}$, where $d_{min}$ and $d_{max}$ denote the minimum and maximum distances between the center and instance contour in $n$ directions of polar coordinate. There are still two issues (see Figure~\ref{V4}~(c)): (1) it does not focus on the centroid location and distributes without rules; (2) plenty of high weights occur around the instance margin. Different from the centerness, centroidness is formulated as:
\begin{eqnarray}
w_{x,y}=\mathbbm{1}_{\{l_{x,y}^*\in mask\}}\frac{\mathrm{min}\{d_{x,y}^1,d_{x,y}^2,...,d_{x,y}^m\}}{d_{x,y}^c+\mathrm{min}\{d_{x,y}^1,d_{x,y}^2,...,d_{x,y}^m\}},
\end{eqnarray}   
where $w_{x,y}$ is the centroid weight at the coordinate of $(x,y)$. $d_{x,y}^c$ is the Euclidean distance between the coordinate of $(x,y)$ and the instance centroid. $\{d_{x,y}^1,d_{x,y}^2,...,d_{x,y}^m\}$ denotes the set of Euclidean distances between the coordinate $(x,y)$ and all $m$ instance contour points. $\mathbbm{1}_{\{l_{x,y}^*\in mask\}}$ is the indicator function, being 1 if the pixel location ($l_{x,y}^*$) within the range of instance masks and 0 otherwise.

\begin{figure}
	\centering
	\includegraphics[width=.48\textwidth]{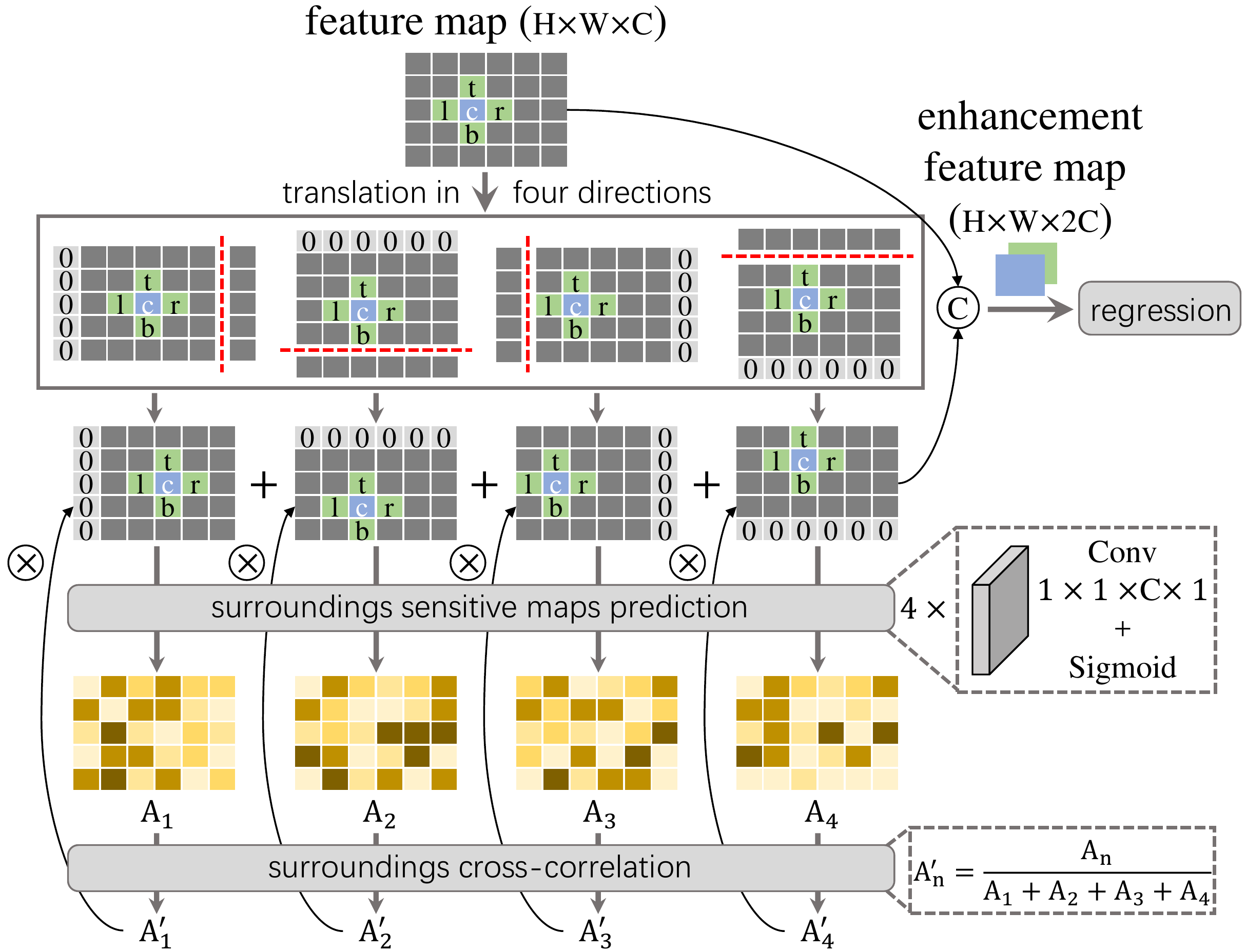}
	\caption{Illustration of surroundings cross-correlation sensitive (SCCS) module structure. $H$, $W$, and $C$ are the height, width, and channel of input feature map. `$\otimes$' and `$\copyright$' are multiplication and concatenation operators, respectively.}
	\label{V5}
\end{figure}

As shown in Figure~\ref{V4}~(d), compared to the centerness proposed in FCOS and PolarMask, we set the centroidness weight to 0 for the background. Meanwhile, the centroidness weight spreads around in a Gaussian distribution related to instance shapes. It makes the weight distribution easier to learn, which encourages our model for retaining high-quality instances more effectively. 

\textbf{Regression head.} The regression head is used to regress the veins (see Figure~\ref{V1}~(a)). As depicted in Figure~\ref{V1}~(d), our method needs to regress minor veins at the node for refining instance twisty parts. However, different from the root, the node is close to the instance contour, which leads to weak features for regressing minor veins. We follow this issue and introduce a surroundings cross-correlation sensitive (SCCS) module to help the vein regression at the node. 

The module aims to enhance feature expression through surrounding information. Considering the enhanced pixel plays a dominant role in the vein regression tasks, we concatenate the pixel feature and the cross-correlated surrounding features instead of concatenating all features. Meanwhile, since the different sensitivity of enhanced pixels to surrounding information in different directions, surroundings sensitive prediction is introduced to help focus on the key information. Specifically, as shown in Figure~\ref{V5}, each feature map from FPN is translated in left, top, right, and bottom directions to generate four translation features at first. Then, the corresponding surroundings sensitive maps are predicted ($\mathrm{A_1, A_2, A_3, A_4}$) and cross-correlated ($\mathrm{A_1^{'}, A_2^{'}, A_3^{'}, A_4^{'}}$). In the end, combining the translation features and surroundings sensitive maps for enhancing the vein regression at the node. 

\begin{figure*}
	\centering
	\includegraphics[width=.9\textwidth]{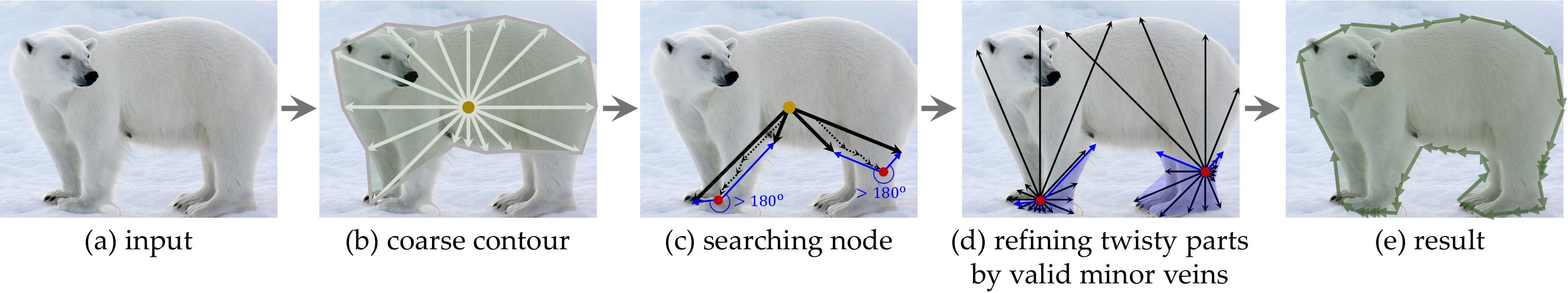}
	\caption{Illustration of vein growth based instance mask assembling process. The `\textbf{white lines}' in (b) are major veins. The `\textbf{black lines}' in (c) are major veins that are used for searching twisty part nodes (`\textbf{red circle}'). The `\textbf{black dot lines}' in (c) are probes. The `\textbf{black lines}' in (d) are minor veins and the lines within the range of `\textbf{blue region}' are valid minor veins that are responsible for refining twisty parts.}
	\label{V6}
\end{figure*}

\textbf{Vein growth.} We introduce a natural and effective representation method (VeinMask) for covering instances masks precisely in low design complexity. It assembles instance contours by simulating the leaf vein growth process. 

Concretely, see Figure~\ref{V6}, given an instance, our method first locates the instance centroid (root) through classification and centroidness heads. Then, the distances between the centroid and instance contour in $n$ directions of the polar coordinate are obtained by the regression head. Combining the centroid and the distances to generate $n$ major veins, where a coarse instance contour can be constructed by connecting the vein endpoints in the clockwise direction (see Figure~\ref{V6}~(b)). It is found that the instance is decomposed into $n$ parts via major veins. Next, all twisty parts are refined by minor veins one by one (see Figure~\ref{V6}~(c)--(d)). For a specific part, our method searches the corresponding node under the guidance of adjacent major veins (the details can be referred to Algorithm~\ref{algorithm1}) and determines it is a twisty part if the angle $<\overrightarrow{c_ne_{pre}^n},\overrightarrow{c_ne_{nxt}^n}>$ is bigger than $180^{\circ}$, where $c_n$ is the node coordinate of $n$th part. $e_{pre}^n$ and $e_{nxt}^n$ denote the coordinates of $n$th part adjacent major vein endpoints, respectively (see Figure~\ref{V6}~(c)). At the location of the twisty part node, the directions in the range of $(\overrightarrow{c_ne_{pre}^n},\overrightarrow{c_ne_{nxt}^n})$ are treated as minor veins and the vein endpoints are connected in the clockwise direction can be obtained the contour points of this twisty part (see Figure~\ref{V6}~(d)).

\begin{algorithm}[t]
	\caption{Search Node} 
	\label{algorithm1} 
	\begin{algorithmic}[1]
		\Require The offset maps $M$; the instance centroid coordinate $c$; searching deep $s$;
		\Ensure  The node coordinate $c_n$ of $n$th part; 
		\For{$i=0 \to s$}
		\If{$i==s-1$}
		\State $\lambda_n^1 \gets \frac{1}{2}$; $\lambda_n^2 \gets \frac{1}{2}$
		\Else
		\State $\lambda_n^1 \gets \frac{1}{2(s-i)-1}$; $\lambda_n^2 \gets \frac{1}{2(s-i)-2}$
		\EndIf
		\State $l_{n} \gets M[c[1], c[0]][n]$
		\State $c_n \gets (\lambda_n^1l_n+c[0],\lambda_n^1l_n+c[1])$
		\State $l_{n+1} \gets M[c_n[1], c_n[0]][n+1]$
		\State $c_n \gets (\lambda_n^2l_{n+1}+c_n[0],\lambda_n^2l_{n+1}+c_n[1])$
		\State $c \gets c_n$
		\EndFor
	\end{algorithmic}  
\end{algorithm}

\textbf{Residual IoU loss.} As depicted in Figure~\ref{V1}, our method simulates the leaf vein growth process to assemble instance contours via the veins. To train them effectively, we introduce Residual IoU (R-IoU) loss.

Concretely, IoU loss~\cite{yu2016unitbox} and the corresponding improved versions~\cite{rezatofighi2019generalized,zheng2020distance,zhang2022focal} are designed for supervising the four bounds of the box as a whole. Considering the IoU of irregular-shaped instances is difficult to compute, Polar IoU loss~\cite{xie2020polarmask} expresses it via the ratio of dense discrete offsets. However, the loss may result in different optimization gradients for the same differences between the predicted and real values, which brings interference to the optimization process. we follow this issue and design R-IoU loss:
\begin{eqnarray}
{\cal L}_{R-IoU}=\frac{\sum_{k=1}^{n}|d_k^{}-d_{k}^*|}{\sum_{k=1}^{n}d_{k}^*}, 
\end{eqnarray}   
where $d_k$ and $d_{k}^*$ are the predicted and real values, respectively. $n$ is the direction number of polar coordinate.

Rather than computing Polar IoU loss via the ratio of the predicted and real value, we explicitly let the model takes the residual over them as the optimization object directly, which avoids the problem that exists in Polar IoU loss effectively and is easier to learn for the model. Moreover, compared with the logarithm formulation of Polar IoU loss, the derivation of our loss function is simpler, which helps improve the back propagation efficiency. 

For classification and centroidness heads, we adopt cross-entropy loss (${\cal L}_{CE}$) and focal loss (${\cal L}_{FL}$)~\cite{lin2017focal} to supervise them. The overall loss ${\cal L}$ can be formulated as:
\begin{eqnarray}
{\cal L}={\cal L}_{R-IoU}+{\cal L}_{FL}+{\cal L}_{CE}.
\end{eqnarray} 

\section{Experiments}
\label{Sec:Experiments}
\textbf{Datasets.} We show the experimental results on the SBD~\cite{hariharan2011semantic} and COCO \textit{test-dev}~\cite{lin2014microsoft} datasets in this section. We typically analyze the superiorities of VeinMask over the other contour-based methods on the SBD dataset in detail. Meanwhile, we compare our approach with existing state-of-the-art (SOTA) methods on the two datasets.

\textbf{Training details.} In ablation study, ResNet-50~\cite{he2016deep} pre-trained on ImageNet~\cite{deng2009imagenet} is adopted as backbone unless otherwise noted. We resize the input into 768$\times$768. The model is trained by the optimizer of stochastic gradient descent (SGD) for 144 epochs on the SBD dataset, where weight decay and momentum are set as 0.0001 and 0.9. The initial learning rate and batch size are 0.01 and 16. In comparison experiments, the images on the COCO \textit{test-dev} dataset are scaled to 800$\times$1333 and padded to 896$\times$1408 with 0.

\subsection{Ablation Study}
\begin{figure}
	\centering
	\includegraphics[width=.4\textwidth]{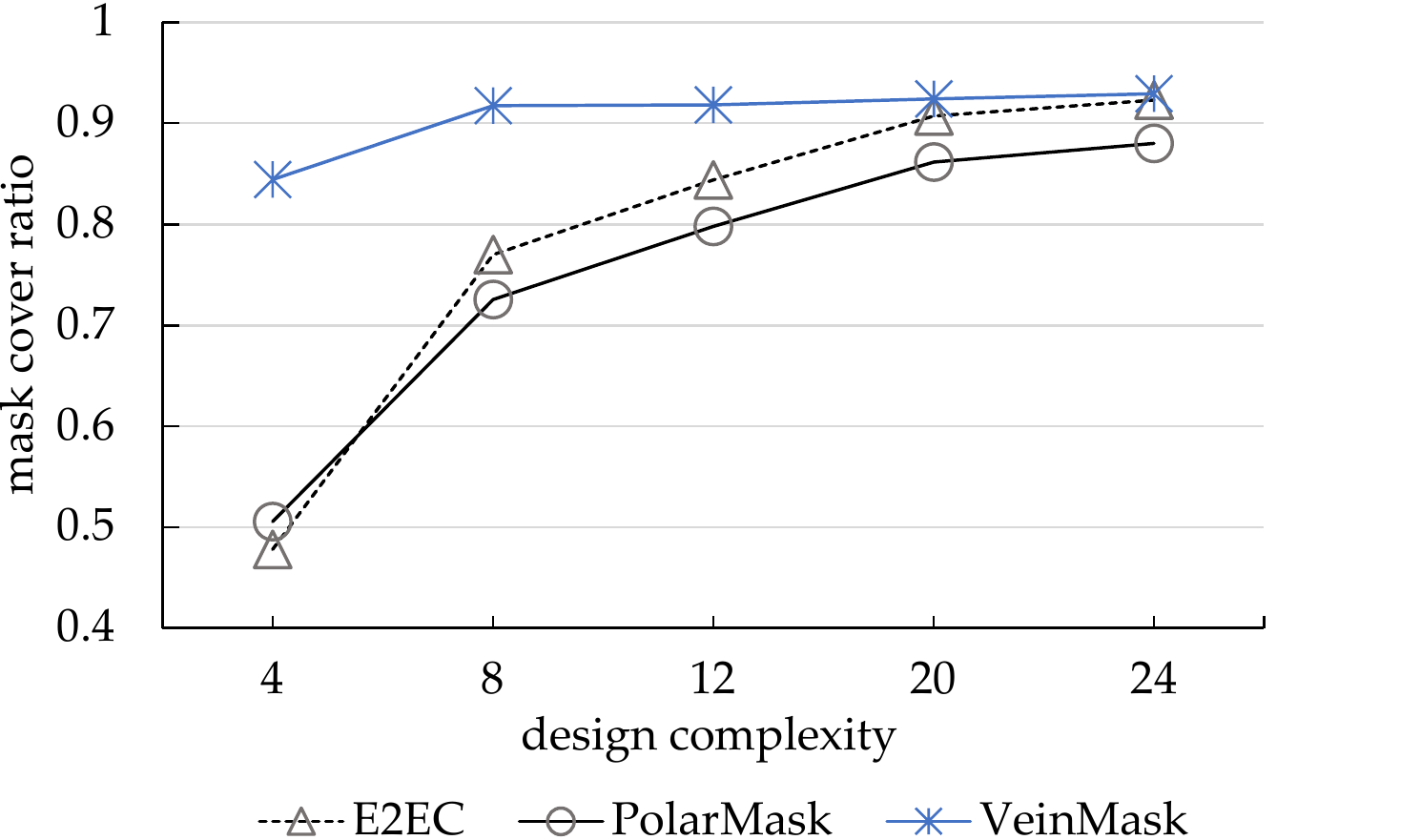}
	\caption{\textbf{Mask cover ratio.} The mask cover ratio upper bound analysis of different methods and different design complexities.}
	\label{V7}
\end{figure}

We ablate our approach in Figure~\ref{V7} and Table~\ref{T1} to verify the effectiveness of the proposed VeinMask, centroidness, SCCS, and R-IoU loss. The design complexity is set to 12 unless otherwise noted. The default model in Table~\ref{T1} is constructed with the centerness and polar IoU loss proposed in PolarMask~\cite{xie2020polarmask} and without VeinMask and SCCS.

\begin{table*}[]
	\renewcommand{\arraystretch}{1}
	\setlength{\tabcolsep}{0.8mm}
	\quad\quad
	\begin{subtable}{0.42\linewidth}
		\centering
		\begin{tabular}{c|ccc ccc}
			case & $\mathrm{AP}$ & $\mathrm{AP_{50}}$ & $\mathrm{AP_{75}}$ & $\mathrm{AP}_S$ & $\mathrm{AP}_M$ & $\mathrm{AP}_L$ \\ \Xhline{1pt}
			w/o   &  24.4  &   53.3   &  19.4    &  8.4   &  17.0   &  29.8   \\ 
			w     &\cellcolor{gray!20}  \textbf{27.5}  &\cellcolor{gray!20}   53.7   &\cellcolor{gray!20}   24.3   &\cellcolor{gray!20}   8.4  &\cellcolor{gray!20}  17.7   &\cellcolor{gray!20}  34.6   \\ 
		\end{tabular}
		\caption{\textbf{VeinMask.} The model without VeinMask denotes the `minor vein' will not participate in the inference process.}
		\label{T1a}
	\end{subtable}\quad\quad\quad\quad
	\begin{subtable}{0.42\linewidth}
		\centering
		\begin{tabular}{c|cccccc}
			case & $\mathrm{AP}$ & $\mathrm{AP_{50}}$ & $\mathrm{AP_{75}}$ & $\mathrm{AP}_S$ & $\mathrm{AP}_M$ & $\mathrm{AP}_L$ \\ \Xhline{1pt}
			centerness   &  27.5  &   53.7   &   24.3   &   8.4  &  17.7   &  34.6     \\ 
			centroidness &\cellcolor{gray!20}  \textbf{29.1}  &\cellcolor{gray!20}   56.6   &\cellcolor{gray!20}   26.1   &\cellcolor{gray!20}   7.4  &\cellcolor{gray!20}  19.0   &\cellcolor{gray!20}  36.9   \\ 
		\end{tabular}
		\caption{\textbf{Centroidness \textit{vs}. centerness.} Our centroidness can improve
			the filtering of low-quality instances compared with centerness.} 
		\label{T1b}
	\end{subtable}
	\vspace{2pt}
	
	\quad\quad
	\begin{subtable}{0.42\linewidth}
		\centering
		\begin{tabular}{c|cccccc}
			case & $\mathrm{AP}$ & $\mathrm{AP_{50}}$ & $\mathrm{AP_{75}}$ & $\mathrm{AP}_S$ & $\mathrm{AP}_M$ & $\mathrm{AP}_L$ \\ \Xhline{1pt}
			w/o          &  27.5  &   53.7   &   24.3   &   8.4  &  17.7   &  34.6     \\ 
			w            &\cellcolor{gray!20}  \textbf{29.0}  &\cellcolor{gray!20}   56.6   &\cellcolor{gray!20}   25.8   &\cellcolor{gray!20}   7.7  &\cellcolor{gray!20}  19.3   &\cellcolor{gray!20}  36.8   \\ 
		\end{tabular}
		\caption{\textbf{SCCS.} The regression head with SCCS helps segment instances more accurately.}
		\label{T1c}
	\end{subtable}\quad\quad\quad\quad
	\begin{subtable}{0.42\linewidth}
		\centering
		\begin{tabular}{c|cccccc}
			reg loss & $\mathrm{AP}$ & $\mathrm{AP_{50}}$ & $\mathrm{AP_{75}}$ & $\mathrm{AP}_S$ & $\mathrm{AP}_M$ & $\mathrm{AP}_L$ \\ \Xhline{1pt} 
			Polar IoU      &  27.5  &   53.7   &   24.3   &   8.4  &  17.7   &  34.6     \\ 
			R-IoU          &\cellcolor{gray!20}  \textbf{28.8}  &\cellcolor{gray!20}   55.6   &\cellcolor{gray!20}   26.0   &\cellcolor{gray!20}   6.7  &\cellcolor{gray!20}  18.0   &\cellcolor{gray!20}  37.3   \\ 
		\end{tabular}
		\caption{\textbf{R-IoU loss \textit{vs.} Polar IoU loss.} R-IoU loss encourages the model to learn strong expression instance features.}
		\label{T1d}
	\end{subtable}
	\vspace{2pt}
	
	\quad\quad
	\begin{subtable}{0.42\linewidth}
		\centering
		\begin{tabular}{c|cccccc}
			complexity & $\mathrm{AP}$ & $\mathrm{AP_{50}}$ & $\mathrm{AP_{75}}$ & $\mathrm{AP}_S$ & $\mathrm{AP}_M$ & $\mathrm{AP}_L$ \\ \Xhline{1pt}
			4   &  21.1  &   50.7   &   15.2   &  1.2   &  14.1   &  28.7   \\ 
			8   &  28.6  &   56.0   &   25.3   &  5.9   &  19.7   &  36.1   \\ 
			12  &  29.4  &   56.4   &   26.8   &  7.8   &  20.0   &  37.0   \\ 
			20  &  30.5  &   58.2   &   27.5   &  8.3   &  20.8   &  38.0   \\ 
			24  &\cellcolor{gray!20}  \textbf{30.7}  &\cellcolor{gray!20}   58.1   &\cellcolor{gray!20}   28.0   &\cellcolor{gray!20}  9.3   &\cellcolor{gray!20}  21.5   &\cellcolor{gray!20}  38.2   \\ 
		\end{tabular}
		\caption{\textbf{Design complexity.} More directions in polar coordinates bring a large gain, while too many directions saturate.}
		\label{T1e}
	\end{subtable}\quad\quad\quad\quad
	\begin{subtable}{0.42\linewidth}
		\centering
		\begin{tabular}{c|cccccc}
			deep & $\mathrm{AP}$ & $\mathrm{AP_{50}}$ & $\mathrm{AP_{75}}$ & $\mathrm{AP}_S$ & $\mathrm{AP}_M$ & $\mathrm{AP}_L$ \\ \Xhline{1pt}
			none  &  25.9  &   55.2   &   21.3   &  7.8   &  17.9   &  32.1   \\ 
			+P7   &  26.0  &   55.3   &   21.3   &  7.8   &  17.9   &     32.2\\  
			+P6   &  28.3  &   55.8   &   25.0   &  7.8   &  18.0   &     36.2\\  
			+P5   &  29.3  &   56.3   &   26.8   &  7.8   &  19.2   &     37.4\\  
			+P4   &\cellcolor{gray!20}  \textbf{29.4}  &\cellcolor{gray!20}   56.4   &\cellcolor{gray!20}   26.8   &\cellcolor{gray!20}  7.8   &\cellcolor{gray!20}  20.0   &\cellcolor{gray!20}  37.0     \\  
			+P3   &  29.1  &   56.1   &   26.3   &  6.7   &  19.1   &     37.4\\  
		\end{tabular}
		\caption{\textbf{Deep for refining the twisty part.} Growing minor veins on more different-sized feature maps brings more gains except on P3.}
		\label{T1f}
	\end{subtable}
	\caption{\textbf{Ablation experiments on the SBD.} We report the effectiveness of VeinMask, centroidness, SCCS, and R-IoU loss in (a)--(d), respectively. Furthermore, we explore the performance of the model with different design complexities, and the model that refining twisty parts via minor veins on different feature maps.}
	\label{T1}
\end{table*}

\textbf{Mask cover ratio.} VeinMask is designed to cover instance masks precisely with low design complexity. We compare the mask cover ratio upper bound of it with previous representative contour-based methods (PolarMask~\cite{xie2020polarmask} and E2EC~\cite{zhang2022e2ec}) in different design complexities. Notably, the design complexity denotes the direction number of the polar coordinate in our method and PolarMask, and the vertices number for E2EC. As depicted in Figure~\ref{V7}, benefiting from the advantages of the dynamic fitting process, VeinMask achieves almost 92\% mask cover ratio with a design complexity of 8, which outperforms the others a lot. Particularly, compared with the same type method (PolarMask), VeinMask is still in a dominant position even when tuning design complexity to large. The results verify the VeinMask's superior ability to cover masks. 

\textbf{VeinMask.} The effectiveness of VeinMask is studied in Table~\ref{T1a}. It is found that refining twisty parts via the minor vein of VeinMask can bring 3.1\% improvements in mAP, which verifies the VeinMask's strong ability to segment instance masks with complex shapes.

\begin{figure}
	\centering
	\subcaptionbox{Loss curve of polar mask}{\includegraphics[width=.47\linewidth]{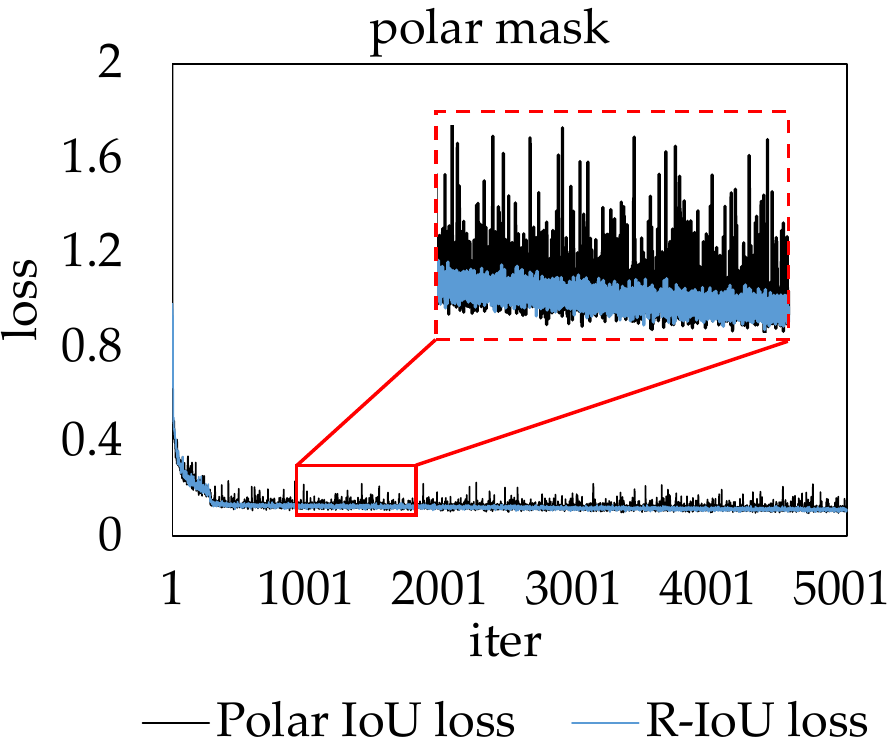}}
	\subcaptionbox{PR curve of polar mask}{\includegraphics[width=.47\linewidth]{fig/V8_c.pdf}}
	\caption{\textbf{Training process.} R-IoU loss enjoys more efficient training process than Polar IoU loss.}
	\label{V8}
\end{figure}

\textbf{Centroidness \textit{vs}. centerness.} An important design of contour-based approaches is to filter low-quality results via centerness. In this work, we propose centroidness for instance segmentation tasks according to the relation between instance geometric characteristics and centroid distribution. 

\begin{table}[]
	\centering
	\renewcommand{\arraystretch}{1}
	\setlength{\tabcolsep}{0.8mm}
	\begin{tabular}{c|cccccc}
		model        & $\mathrm{AP}$ & $\mathrm{AP_{50}}$ & $\mathrm{AP_{75}}$ & $\mathrm{AP}_S$ & $\mathrm{AP}_M$ & $\mathrm{AP}_L$ \\ \Xhline{1pt} 
		PolarMask~\cite{xie2020polarmask} &  25.9  &    57.0  &   20.3   &  8.0   &  18.4   &  32.0   \\ 
		with R-IoU loss   &  27.8  &   59.1   &   22.3   &  8.8   &  19.7   &  34.4   \\ 
		with centroidness &  28.3  &   59.9   &   22.6   &  8.4   &  19.8   &  35.0   \\ 
	\end{tabular}
	\caption{\textbf{Generalization of centroidness and R-IoU loss.} We replace the centerness and Polar IoU loss in PolarMask via centroidness and R-IoU loss to show the brought performance gains.}
	\label{T2}
\end{table}

Table~\ref{T1b} studies this design. We report the performance of models that are equipped with centerness and centroidness. It is found that centroidness brings 1.6\% mAP for our model, which benefits from the advantage that centroidness ensures the result's reliability in the inference process. The results verify the superior of centroidness over centerness.

\textbf{SCCS.} We design SCCS module to enhance feature expression to help regress veins more accurately. As shown in Table~\ref{T1c}, the module brings 1.5\% mAP improvements to the baseline model. Rather than extracting features of all contour points and concatenating them, SCCS ensures that the regression head and the other two heads run in parallel. It avoids multiple stages serial feature enhancement process in~\cite{peng2020deep,zhang2022e2ec}, which brings gains with negligible costs.

\begin{figure*}
	\centering
	\subcaptionbox{w/o VeinMask}{\includegraphics[width=.15\linewidth]{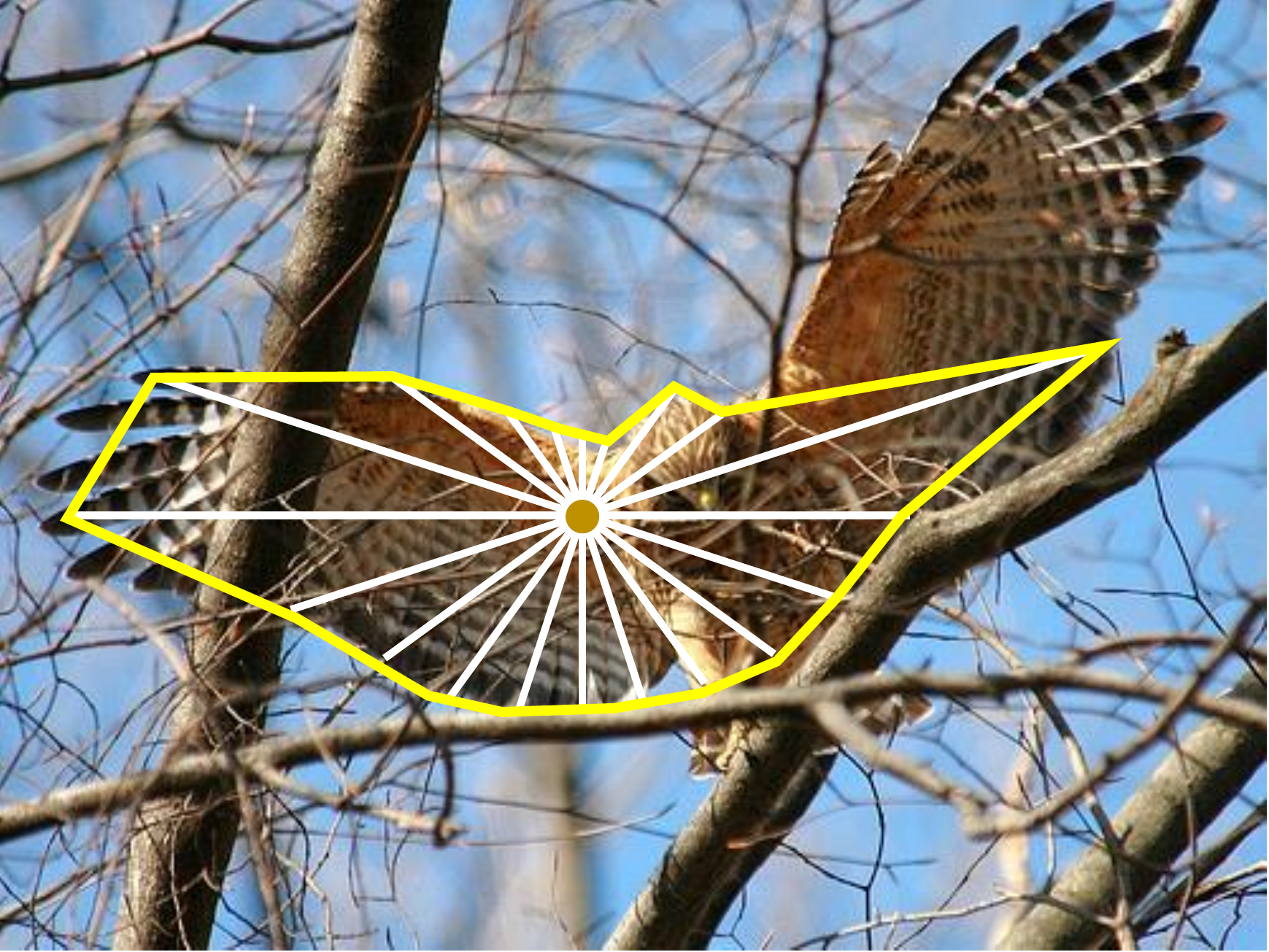}}
	\subcaptionbox{with VeinMask}{\includegraphics[width=.15\linewidth]{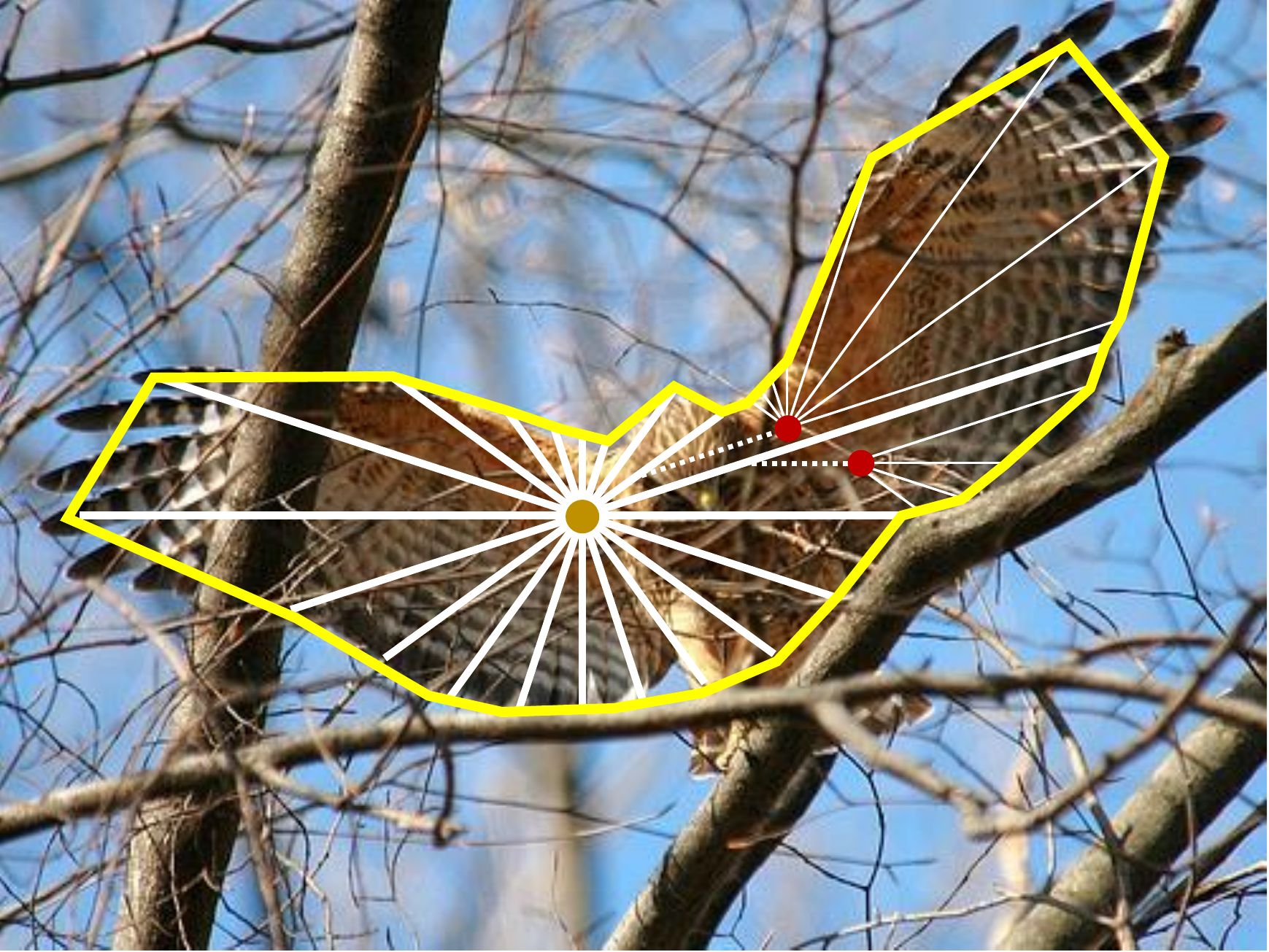}}
	\subcaptionbox{result in SBD}{\includegraphics[width=.15\linewidth]{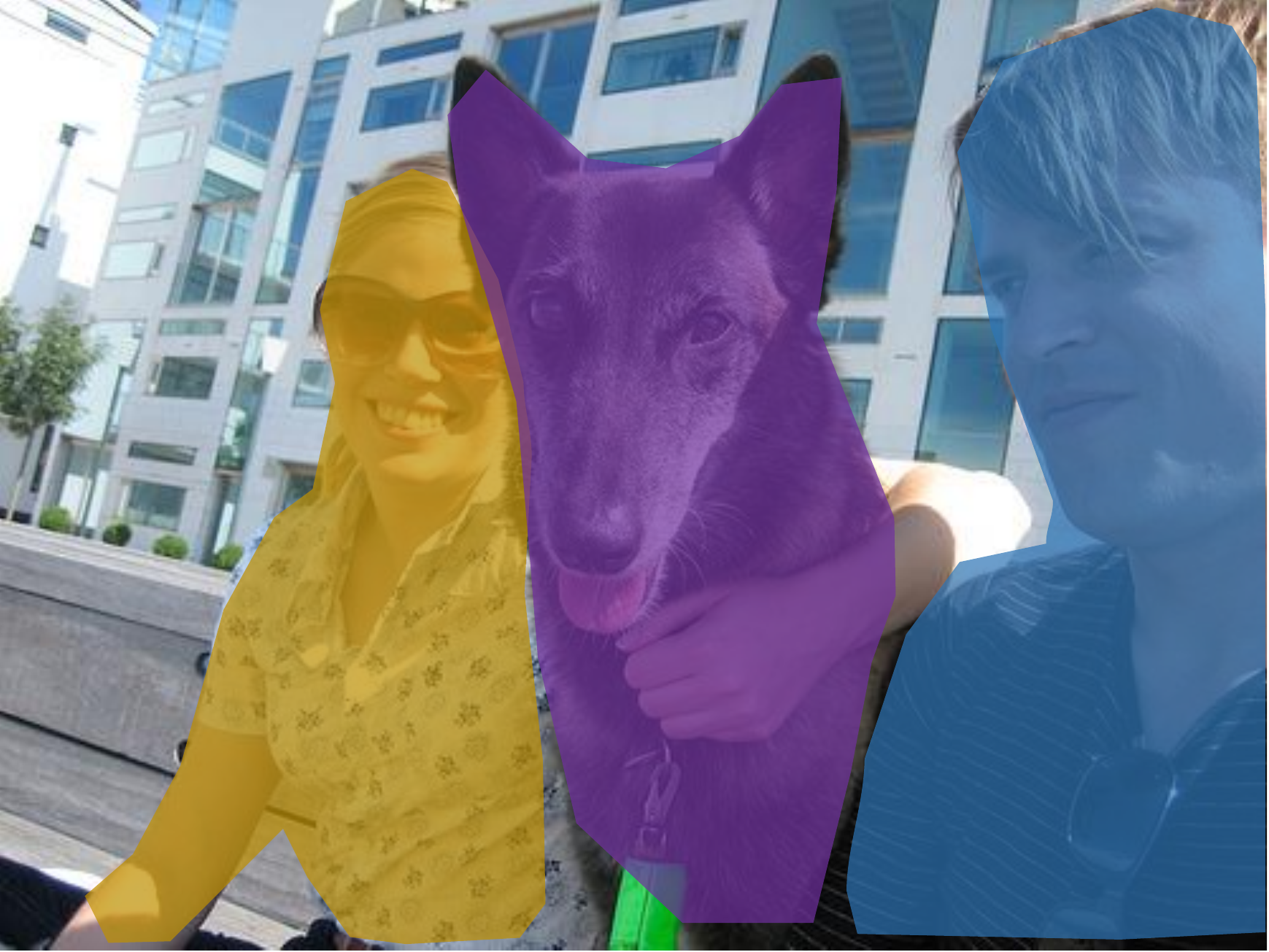}}
	\subcaptionbox{result in SBD}{\includegraphics[width=.15\linewidth]{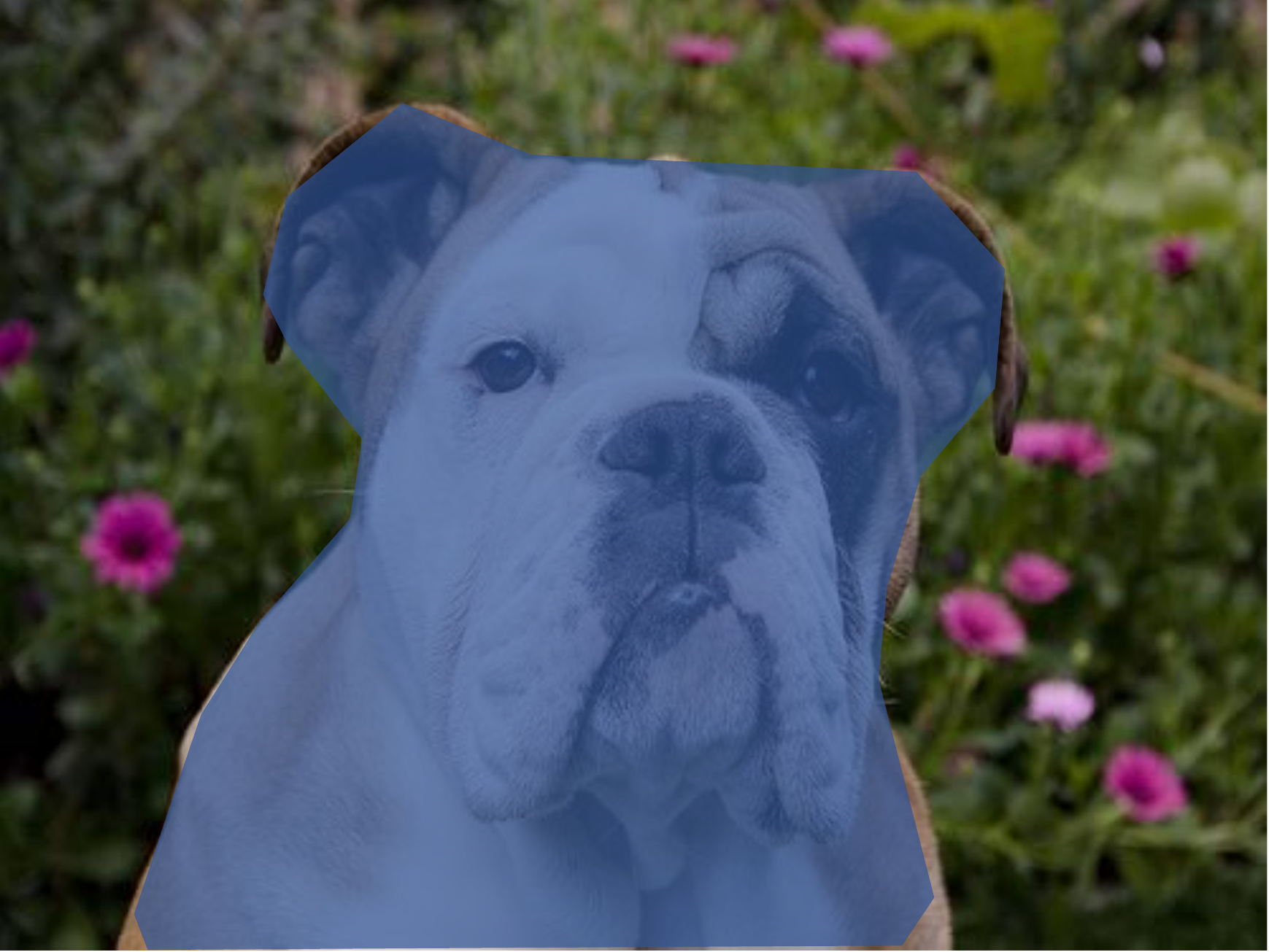}}
	\subcaptionbox{result in COCO}{\includegraphics[width=.15\linewidth]{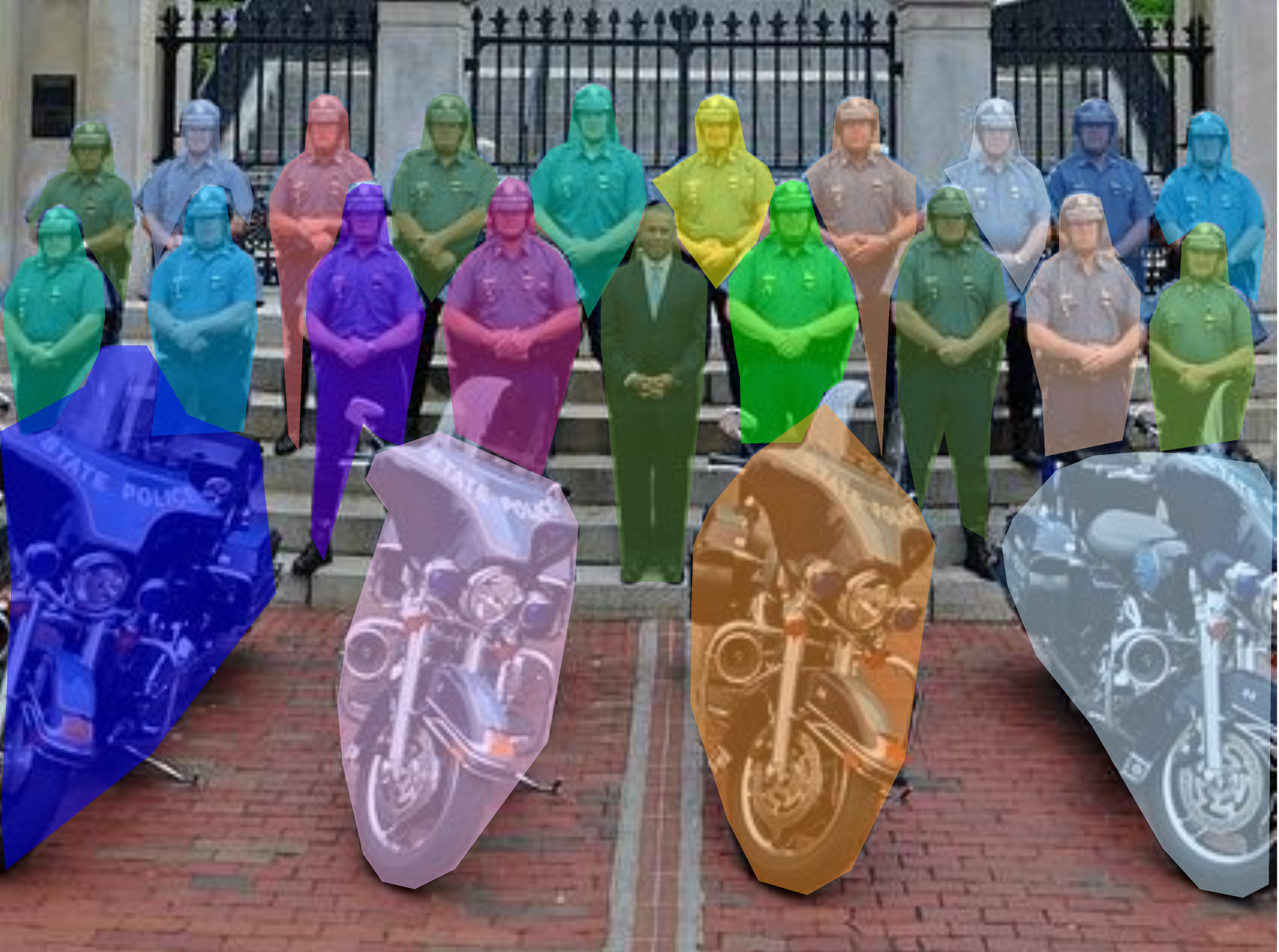}}
	\subcaptionbox{result in COCO}{\includegraphics[width=.15\linewidth]{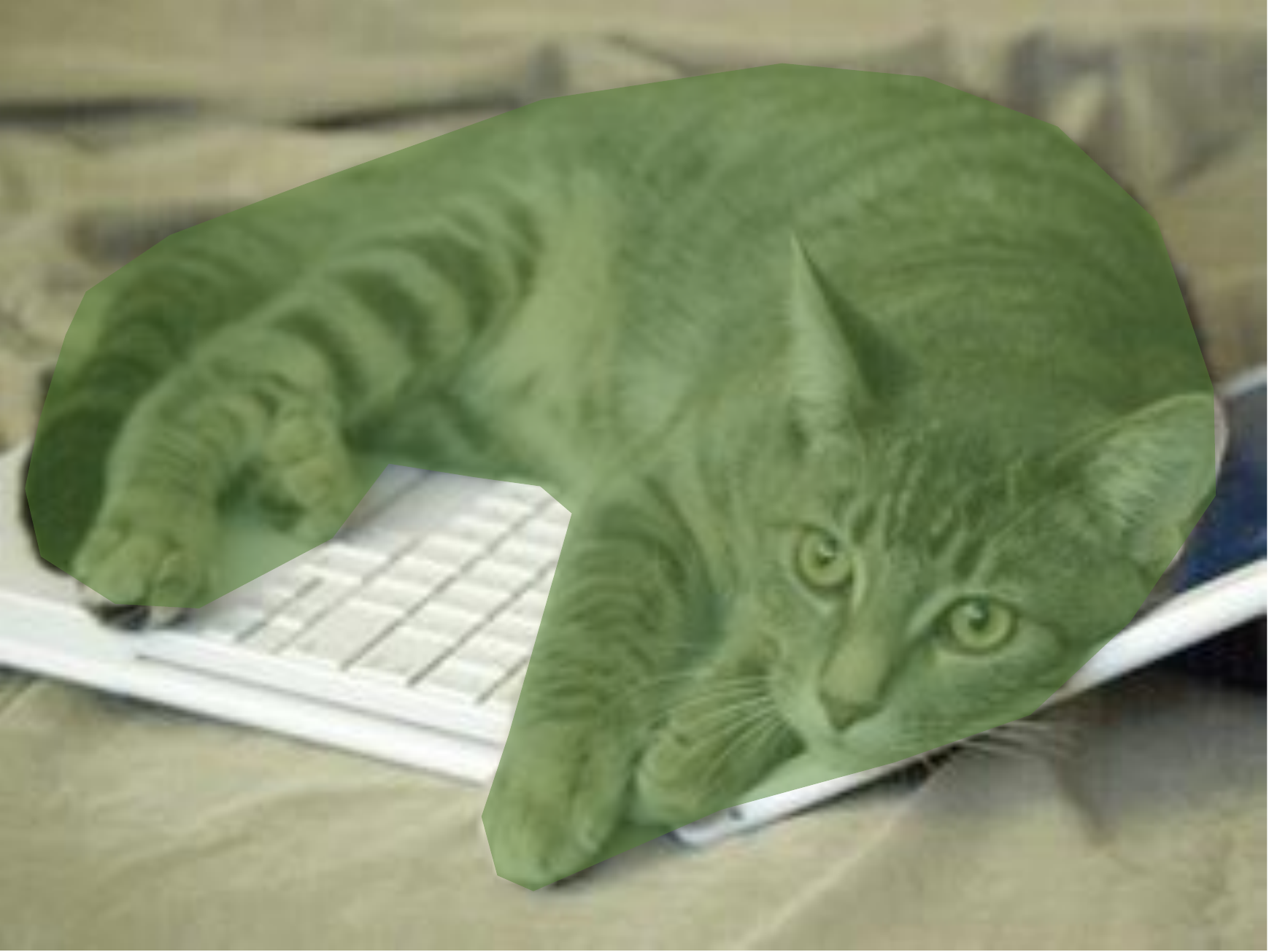}}
	\caption{Visualization of some qualitative results of VeinMask on the SBD and COCO \textit{test-dev} datasets.}
	\label{V9}
\end{figure*}

\begin{table*}[]
	\begin{subtable}{0.46\linewidth}
		\centering
		\renewcommand{\arraystretch}{1.2}
		\setlength{\tabcolsep}{0.8mm}
		\begin{tabular}{l|c|ccc|ccc}
			\multirow{2}{*}{method} & \multirow{2}{*}{c} & \multicolumn{3}{c|}{SBD}                                     & \multicolumn{3}{c}{COCO}                                    \\ \cline{3-8} 
			&                             & \multicolumn{1}{c}{$\mathrm{AP}$}   & \multicolumn{1}{c}{$\mathrm{AP_{50}}$} & $\mathrm{AP_{75}}$ & \multicolumn{1}{c}{$\mathrm{AP}$}   & \multicolumn{1}{c}{$\mathrm{AP_{50}}$} & $\mathrm{AP_{75}}$ \\ \Xhline{1pt}
			PolarMask~\cite{xie2020polarmask}               & 4                           & \multicolumn{1}{c}{5.1}  & \multicolumn{1}{c}{21.1} & 0.1  & \multicolumn{1}{c}{1.1}  & \multicolumn{1}{c}{6.2}  & 0.0  \\ 
			E2EC~\cite{zhang2022e2ec}                    & 4                           & \multicolumn{1}{c}{2.3}  & \multicolumn{1}{c}{16.4} & 0.0  & \multicolumn{1}{c}{2.0}  & \multicolumn{1}{c}{11.5} & 0.0  \\ 
			VeinMask                & 4                           & \multicolumn{1}{c}{21.1} & \multicolumn{1}{c}{50.7} & 15.2 & \multicolumn{1}{c}{14.8} & \multicolumn{1}{c}{34.2} & 11.8 \\ \Xhline{1pt}
			PolarMask~\cite{xie2020polarmask}               & 8                           & \multicolumn{1}{c}{20.7} & \multicolumn{1}{c}{52.5} & 12.9 & \multicolumn{1}{c}{5.6}  & \multicolumn{1}{c}{14.0} & 3.2  \\ 
			E2EC~\cite{zhang2022e2ec}                    & 8                           & \multicolumn{1}{c}{26.5} & \multicolumn{1}{c}{57.6} & 20.7 & \multicolumn{1}{c}{17.3} & \multicolumn{1}{c}{37.4} & 14.3 \\ 
			\textbf{VeinMask}                & 8                           & \multicolumn{1}{c}{\cellcolor{gray!20}\textbf{28.6}} & \multicolumn{1}{c}{\cellcolor{gray!20}56.0} &\cellcolor{gray!20} 25.3 & \multicolumn{1}{c}{\cellcolor{gray!20}\textbf{25.3}} & \multicolumn{1}{c}{\cellcolor{gray!20}47.5} &\cellcolor{gray!20} 24.0 \\ 
		\end{tabular}
		\caption{\textbf{Comparisons with contour-based methods on the SBD and COCO \textit{test-dev}.} VeinMask can cover masks precisely in low design complexity and the corresponding performance outperforms the others a lot.}
		\label{T3a}
	\end{subtable}\quad\quad\quad
	\begin{subtable}{0.46\linewidth}
		\centering
		\renewcommand{\arraystretch}{1}
		\setlength{\tabcolsep}{0.8mm}
		\begin{tabular}{l|l|c|ccc}
			type                           & method      & c & $\mathrm{AP}$   & $\mathrm{AP_{50}}$ & $\mathrm{AP_{75}}$  \\ \Xhline{1pt}
			\multirow{6}{*}{\textit{{mask-based}}}    & \textit{two-stage} &            &      &      &      \\  
			& Mask R-CNN~\cite{he2017mask}    & -          & 35.7 & 58.0 & 37.8  \\
			& BoundFormer~\cite{lazarow2022instance}    & -         & 37.7 & 58.8 & 40.5  \\ \cline{2-6} 
			& \textit{one-stage}   &          &            &      &    \\ 
			& YOLACT~\cite{bolya2019yolact}   & -          & 31.2 & 50.6 & 32.8  \\ 
			& SOLO~\cite{wang2020solo}        & -          & 37.8 & 59.5 & 40.4 \\ \Xhline{1pt}
			\multirow{7}{*}{\textit{{contour-based}}} & \textit{multi-stage} &            &      &      &      \\ 
			& DeepSnake~\cite{peng2020deep}   & 128        & 30.3 & -    & -    \\ 
			& E2EC~\cite{zhang2022e2ec}       & 128        & 31.7 & 52.2 & 32.8 \\ \cline{2-6} 
			& \textit{one-stage}   &          &            &      &      \\  
			& PolarMask~\cite{xie2020polarmask}   & 36     & 30.4 & 51.9 & 31.0\\  
			& \textbf{VeinMask}    & 20       &  32.4    &   55.7   &   32.7   \\
		\end{tabular}
		\caption{\textbf{Comparisons with all kinds of sota methods on COCO \textit{text-dev}.} VeinMask achieves comparable performance with mask-based methods.} 
		\label{T3b}
	\end{subtable}
	\caption{\textbf{Comparisons with previous results on the SBD and COCO \textit{test-dev} datasets.} `c' denotes design complexity, which denotes the direction number of the polar coordinate in our method and PolarMask, and the vertices number for DeepSnake and E2EC.}
	\label{T3}
\end{table*}

\textbf{R-IoU loss \textit{vs.} Polar IoU loss.} In Table~\ref{T1d} we compare our R-IoU loss with Polar IoU loss. 

We force the model to focus on the residual between the predicted and real values via R-IoU loss, a more effective optimization object than Polar IoU loss. It enjoys a simpler derivative formulation and ensures the same gradients for the same residuals. The experimental results in Table~\ref{T1d} show that our R-IoU loss brings 1.3\% mAP gains, which verifies its superiority over Polar IoU loss. We report the training processes of PolarMask equipped with Polar IoU and R-IoU losses respectively in Figure~\ref{V8}. R-IoU loss enjoys a smoother descending process than Polar IoU loss and the R-IoU loss' PR curve encloses the Polar IoU loss's curve, which shows the superiority of R-IoU loss.

\textbf{Design complexity.} Tuning design complexity to larger brings performance gains for contour-based instance segmentation methods. As illustrated in Table~\ref{T1e}, our method gains a lot of performance improvements when tuning the model design complexity from 4 to 20, while the design complexity of 24 saturates since it enjoys the almost same ability as the design complexity of 20 to depict the instance mask ground-truth (see Figure~\ref{V7}).

\textbf{Deep for refining the twisty part.} VeinMask helps improve performance by refining twisty parts via minor veins (see Table~\ref{T1a}). We further explore the performance gains when growing minor veins on feature maps with different sizes. In Table~\ref{T1f}, we grow the minor veins from P7 to P3, it is found that our strategy brings significant performance gain for large instances, while performs not well for small ones since they are very sensitive to the predicted veins.

\textbf{Generalization of centroidness and R-IoU loss.} Centroidness and R-IoU loss are embedded into PolarMask to replace the centerness and Polar IoU loss. In Table~\ref{T2}, they can bring 2.4\% and 1.9\% mAP gains, respectively. In particular, centroidness and R-IoU loss gain performance significantly for large-scale instances because centroidness helps suppress low-quality results more effectively and R-IoU loss makes the model learn more differences between the predicted and real values. The results show their generalization and verify that they can be embedded into other methods seamlessly to gain performance without costs.

\subsection{Comparisons with Previous Results} 
In Table~\ref{T3a}, we compare our method with existing contour-based methods to show superior performance. Specifically, VeinMask outperforms PolarMask~\cite{xie2020polarmask} 19.7\% mAP on the COCO \textit{test-dev} when tuning the design complexity to 8 and outperforms E2EC~\cite{zhang2022e2ec} 18.8\% mAP on the SBD when tuning the design complexity to 4, which demonstrates the strong ability of VeinMask to segment instance masks in low design complexity.

Furthermore, we explore the upper performance of our algorithm on the COCO \textit{test-dev} and compare it to existing SOTA approaches. As depicted in Table~\ref{T3b}, for multi-stage contour-based methods (DeepSnake~\cite{peng2020deep} and E2EC~\cite{zhang2022e2ec}, the proposed method can achieve 32.4\% mAP with the one-sixth design complexity of them. For the one-stage contour-based method (PolarMask~\cite{xie2020polarmask}), our approach even outperforms it 2\% in mAP with almost half the design complexity, which is comparable
with mask-based methods. Additionally, we visualize some qualitative results in Figure~\ref{V9}.

\section{Discussion and Conclusion}
\label{Sec:Conclusion}
Unifying object detection and instance segmentation problems is a valuable exploration direction. In this work, we follow the intrinsic idea of box detectors to segment instances with center classification and offset regression. We observe leaf vein growth mode and design VeinMask--- an instance representation method used to regress curved offsets between instance centroids and twisty parts with a series of straight offsets dynamically---to provide a natural and intuitive solution for the single-shot anchor-free instance segmentation. Remarkably, the dynamic fitting process brings an extra advantage to our method that covers instance masks precisely with low design complexity, which is an interesting idea and makes it possible to construct a novel basic framework for contour-based methods.

Furthermore, we introduce centroidness and R-IoU loss for enhancing the model's ability to segment instances without costs. Importantly, they can be used directly in existing state-of-the-art center classification and offset regression-based architectures. The performance can be effectively enhanced by replacing their centerness and loss function with our centroidness and R-IoU loss. Additionally, we construct SCCS to help feature expression of the offset regression. It can be embedded into other architectures and brings performance gains with slight computational costs. We hope the idea that combines leaf vein morphology with instance geometric characteristics will inspire future work, and the proposed centroidness, R-IoU loss, and SCCS can become basic components of other approaches.

\bibliographystyle{IEEEtran}
\bibliography{egbib}

\vfill
\end{document}